\documentclass[conference]{IEEEtran}
\IEEEoverridecommandlockouts
\usepackage{cite}
\usepackage{amsmath,amssymb,amsfonts}
\usepackage{algorithmic}
\usepackage{graphicx}
\usepackage{textcomp}
\usepackage{xcolor}
\usepackage[acronym]{glossaries}
\usepackage{float}

\def\BibTeX{{\rm B\kern-.05em{\sc i\kern-.025em b}\kern-.08em
    T\kern-.1667em\lower.7ex\hbox{E}\kern-.125emX}}
\begin{document}


\newacronym{inr}{INR}{implicit neural representation}

\title{Implicit Neural Representations for Deconvolving SAS Images}

\author{Albert Reed$^1$, Thomas Blanford$^{2}$, Daniel C. Brown$^{2, 3}$, Suren Jayasuriya$^{1}$ \\
School of Electrical, Computer and Energy Engineering; Arizona State University$^1$\\
	The Applied Research Laboratory$^2$, Graduate Program in Acoustics$^3$; Pennsylvania State University }

\maketitle

\begin{abstract}
    Synthetic aperture sonar (SAS) image resolution is constrained by waveform bandwidth and array geometry. Specifically, the waveform bandwidth determines a point spread function (PSF) that blurs the locations of point scatterers in the scene. In theory, deconvolving the reconstructed SAS image with the scene PSF restores the original distribution of scatterers and yields sharper reconstructions. However, deconvolution is an ill-posed operation that is highly sensitive to noise. In this work, we leverage implicit neural representations (INRs), shown to be strong priors for the natural image space, to deconvolve SAS images. Importantly, our method does not require training data, as we perform our deconvolution through an analysis-by-synthesis optimization in a self-supervised fashion. We validate our method on simulated SAS data created with a point scattering model and real data captured with an in-air circular SAS. This work is an important first step towards applying neural networks for SAS image deconvolution.
\end{abstract}

\section{Introduction}

\begin{figure*}[t]
\centering
\includegraphics[trim={0cm 10cm 4cm 0cm}, clip, width=0.90\textwidth]{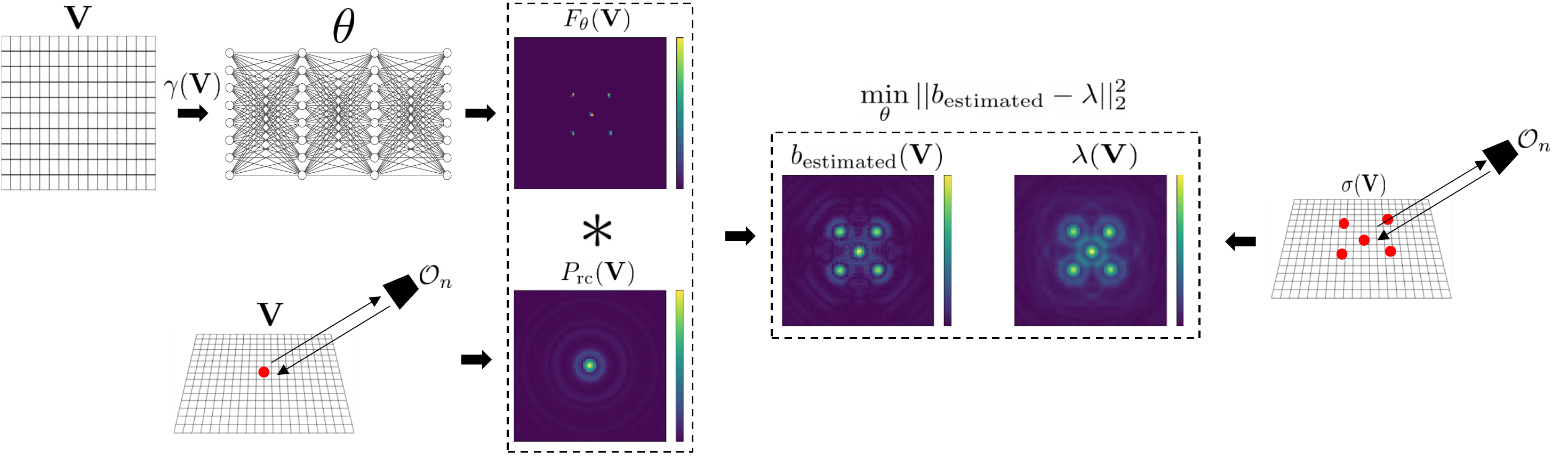}
\caption{We estimate a delay-and-sum reconstructed SAS image by convolving the array PSF with an estimate of the point scattering distribution. We compare estimated measurements with given measurements to compute a loss and iteratively update the network weights until convergence.} 
\label{fig:pipeline}
\end{figure*}

\begin{figure*}[ht!]
\centering
\includegraphics[trim={0cm 7cm 5cm 0cm}, clip, width=0.90\textwidth]{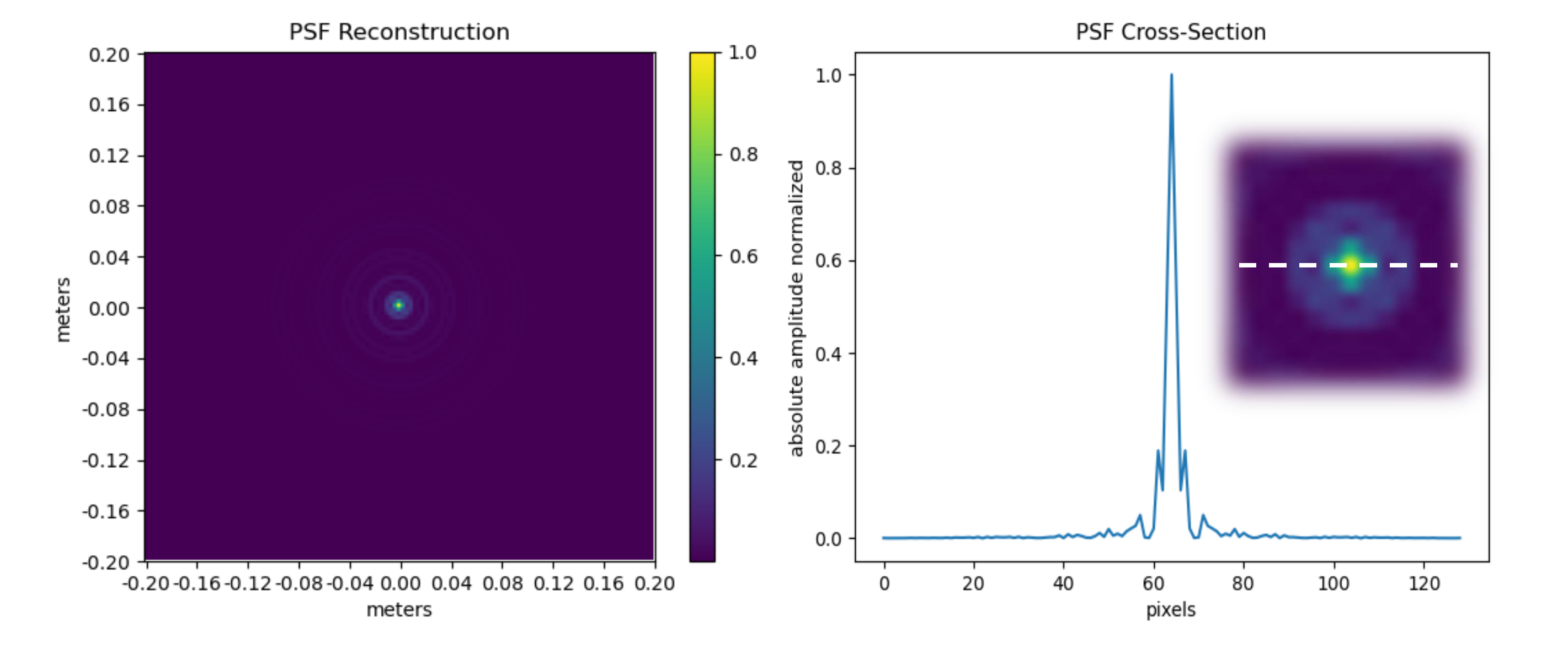}
\caption{Plot of the PSF used in our experiments. We generate this PSF by replica-correlating a $20$kHz LFM and backprojecting the measurements onto the center of a $0.4 \times 0.4$ meter scene, which is the geometry used in our experiments. We observe that the scatterer introduces non-zero amplitudes to a radial neighborhood of approximately 10 pixels.} 
\label{fig:psf_fig}
\end{figure*}

Synthetic aperture sonars (SAS) measure acoustic backscattering by successively pinging a scene from a moving platform. Image reconstruction algorithms coherently combine these measurements to reconstruct high-resolution images of the scene. In practice, SAS operators carefully consider important parameters like waveform design and hardware limitations to maximize image resolution at the desired area coverage rate. In this paper, we focus our attention to the bandwidth of the transmitted waveform and its impacts on image resolution. 

Array geometry and waveform parameters determine a \textit{point spread function} (PSF) that upper bounds the fidelity of reconstructed imagery. Specifically, if we model the measured scene as a distribution of real-valued point scatterers, then the convolution of this PSF with the scatterers describes the measured scene reflectivity function. When the scale of the scene features are smaller than the bandwidth-limited resolution, this convolution effectively smears the location of point scatterers resulting in blurred imagery. However, because the PSF for a particular array geometry and waveform parameters is known, it is theoretically possible to restore the original distribution of scatterers by deconvolving the measured scene with the PSF \cite{de1994deconvolution, pailhas2017increasing, hayes1992broad}. Unfortunately, deconvolution is an ill-posed inverse problem, and algorithms applied to this problem are highly sensitive to noise. Incorporating prior information about the image (e.g., sparsity in some basis) improves results \cite{ren2020neural, bretthorst1992bayesian}, but defining these priors is challenging and often domain specific.  

Our key insight is to leverage recent advances in optical image reconstructions where neural networks learn a function that maps scene coordinates to physical values. Specifically, we use a new type of neural network called an \acrlong{inr} (\acrshort{inr}) that learns a representation of the scene within the network weights \cite{sun2021coil, mildenhall2020nerf}.  This strategy is effective because neural networks are natural priors for the image space. In particular, neural networks trained to learn a scene function demonstrate an affinity for reconstructing meaningful scene features while being robust to noise. These methods are typically self-supervised and do not require training datasets. This phonemenon is termed the \textit{deep image prior} in the computer vision literature \cite{ulyanov2018deep}. 

\textbf{Primary contributions:} We formally outline this paper's contribution as follows: First, we present an algorithm for performing deconvolution to reconstruct SAS images. To deconvolve SAS images, we define an analysis-by-synthesis optimization (i.e., our method does \textit{not} require training data) that optimizes neural network weights to model a function that maps point scattering positions $(x, y)$ to real-valued amplitudes. We validate our method on simulated SAS data generated with a ray-based scattering model and on real data captured from an in-air circular SAS system, called AirSAS \cite{blanford2019development}. We provide a comparison of our reconstructions to delay-and-sum beamformed images, and show that our method reconstructs more detailed image features by deconvolving side lobes from the transmitted waveform. To the best of our knowledge, our work is the first to consider the use of neural networks to deconvolve SAS images. 


\section{Related Work}
\textbf{SAS image reconstruction:} A variety of algorithms exist for reconstructing SAS images from backscattered measurements. The primary class of reconstruction algorithms operate in either the time-domain or frequency domain \cite{hayes1992broad}. Time-domain algorithms delay the signal from each transmitter-receiver pair based on the two-way propagation time for the spatial location of each pixel \cite{gerg2020gpu}. Frequency-domain algorithms map the Fourier transform of measured waveforms to k-space and invoke an inverse Fourier transform to resolve a scene's spatial reconstruction \cite{ferguson2005application, marston2011coherent}. Under standard SAS operating conditions, the two reconstruction approaches are equivalent. Recent work also explores the applicability of SAS image representations to machine learning algorithms \cite{park2020alternative}. In this paper, we utilize a time-domain delay-and-sum beamformer for all our experiments and describe its function with the term \textit{backprojection}, as it coherently backprojects measured acoustic backscattering to the scene. 

\textbf{Deconvolution in optical domain:} Many algorithms for blind and non-blind deconvolution are fundamental to imaging modalities in the optical domain such as microscopes and telescopes. In blind deconvolution, the deconvolution algorithm must jointly estimate the PSF and deconvolved scene. In non-blind deconvolution, the PSF is assumed known and used to reconstruct a representation of the scene. We characterize our approach as non-blind deconvolution since we compute the PSF from the array geometry and waveform parameters. Deconvolution is an ill-posed operation that is notoriously sensitive to noise sources in the scene. Conventional algorithms attenuate noise by reconstructing the scene under a maximum-likelihood framework  or by filtering noise (e.g., Wiener filter) in the frequency domain \cite{richardson1972bayesian, lucy1974iterative}. Additionally, regularizing deconvolution algorithms by leveraging prior information is known to improve results \cite{daun2006deconvolution, krishnan2011blind, krishnan2009fast}. State-of-the-art deconvolution methods leverage untrained and trained neural networks \cite{ren2020neural, xu2014deep}. Our work is most similar to these approaches, as we leverage a neural network as an image space prior to deconvolve SAS images. 

\textbf{Implicit neural representations:} Recently, coordinate-based neural networks called INRs have found great success in the imaging domain for reconstructing 3D scenes from 2D images. These methods are termed implicit representations as they implicitly learn the structure of a scene by learning a function that maps coordinates (e.g., $(x, y)$) to physical properties in the scene (e.g., the density at $(x, y)$). Many works in the optical domain leverage these networks to achieve impressive reconstructions of 3D scenes using only 2D images as input \cite{mildenhall2020nerf, park2020deformable, liu2020neural}. INRs have also recently been applied to tomographic imaging problems such as computed tomography \cite{reed2021dynamic, sun2021coil}, which inspired our motivation for applying them to SAS imaging.

\textbf{Deconvolving SAS images:} Few works consider the deconvolution of SAS imagery. In \cite{pailhas2017increasing}, the authors derive a PSF that is spatially invariant, making it useful for deconvolution of circular-SAS images. In \cite{de1994deconvolution}, the authors propose an iterative SAS deconvolution algorithm and discuss its convergence properties. Both methods validate their methods on numerically simulated SAS data. We have not found works that apply neural networks to SAS deconvolution. 

\section{Methods}

In this section, we present the forward model for SAS imaging and describe how reconstructed SAS images can be described as a convolution of the array PSF with the scene's reflectivity function. We then describe our method for inverting this forward model which leverages a neural network and an analysis-by-synthesis optimization to reconstruct a deconvolved SAS image.  




\textbf{Imaging model:} For our imaging model, we consider a circular SAS geometry where a ring of transducers form a track around the scene, denoted as $\mathcal{O}$. We note that our method is applicable to other array geometries, but we leave these implementations to future work. The scene contains several point scatterers with real-valued amplitudes. These amplitudes modulate the proportion of incident acoustic energy reflected back towards the transducer. Formally, we describe the scatterers with a reflectivity function $\sigma(x, y)$ that describes a distribution of point scatterers. This function maps the scatterers at each $(x, y)$ location to their amplitude --- our goal is to estimate this function using SAS measurements. As such, each transducer pings the scene with a linear frequency modulated (LFM) waveform and records the reflected energy, denoted as $s_{n}(\tau)$ for $\mathcal{O}_n \in \mathcal{O}$. We replica-correlate $s_{n}(\tau)$ with the transmitted LFM waveform and obtain a complex (i.e., real and imaginary) signal $s_{n}^{rc}(\tau)$, and then backproject these measurements onto the estimated scene, $\lambda$, using a delay-and-sum beamformer,

\begin{equation}
\begin{gathered}
    \lambda(x, y) = \int_{O}s_{n}^{rc}(\tau=t_n(x, y, O_n))
\end{gathered} 
\end{equation} where $\tau$ delays the waveform by the propagation time,

\begin{equation}
\begin{gathered}
    t_n(x, y, O_n) = 2||O_n - (x, y)||_2^2 / c
\end{gathered} 
\end{equation} where $c$ is the speed of sound and the transducer is assumed to be monostatic. We note that $\lambda(x, y)$ is complex valued as we consider coherent processing of SAS data. 

{PSF convolution:} Since the transmitted waveform is of finite duration and bandwidth, it contains side lobes that introduce artifacts to the reconstructed scene. Specifically, these side lobes create ambiguities in the reconstructed scatterer amplitudes since measurements from a single scatterer backproject non-zero amplitudes onto a neighborhood of scene points. We characterize the spatial ambiguities as a PSF that we compute by backprojecting returns from a point scatterer centered in the scene. In Figure \ref{fig:psf_fig}, we show the absolute value of the complex PSF used in our experiments, as well as its 1-D cross section. The convolution of this PSF with $\sigma(x, y)$ approximates the measured reconstruction of the scene. This is an approximation because the true PSF is spatially varying as it moves away from the scene center and closer to the transducers. For our experiments, we assume the PSF is spatially invariant for computation's sake, and observe that this approximation yields acceptable results.



The purpose of our method is to attenuate the reconstruction artifacts caused by side lobe energy in the SAS waveform by deconvolving SAS measurements with the scene PSF. We now describe this procedure formally: First, we define $P_{\text{rc}}$ as a complex-valued PSF for a scene. We obtain this PSF by delay-and-sum beamforming replica-correlated measurements reflected from single point scatterer centered in the scene. The estimated reflectivity function for a distribution of scatterers is then,


\begin{equation}
\begin{gathered}
    b(x, y) = \iint_{-\infty}^{\infty} \sigma(u,v) P_{\text{rc}}(x - u, y-v) \,du\,dv \\
    = \sigma*P_{\text{rc}},
\end{gathered} 
\end{equation}
where $b(x, y)$ is complex valued and the SAS image pixel value at each $(x, y)$. Deconvolution restores the original distribution of scatterers, $\sigma = \mathit{IFFT}(\hat{b} \: /\ \hat{P}_{\text{rc}})$ where $( \; \hat{} \;)$ is the Fourier transform operator and $\mathit{IFFT}$ is the inverse Fourier transform \cite{de1994deconvolution}.

\begin{figure*}[ht!]
\centering
\includegraphics[trim={0cm 4cm 0cm 0cm}, clip, width=0.90\textwidth]{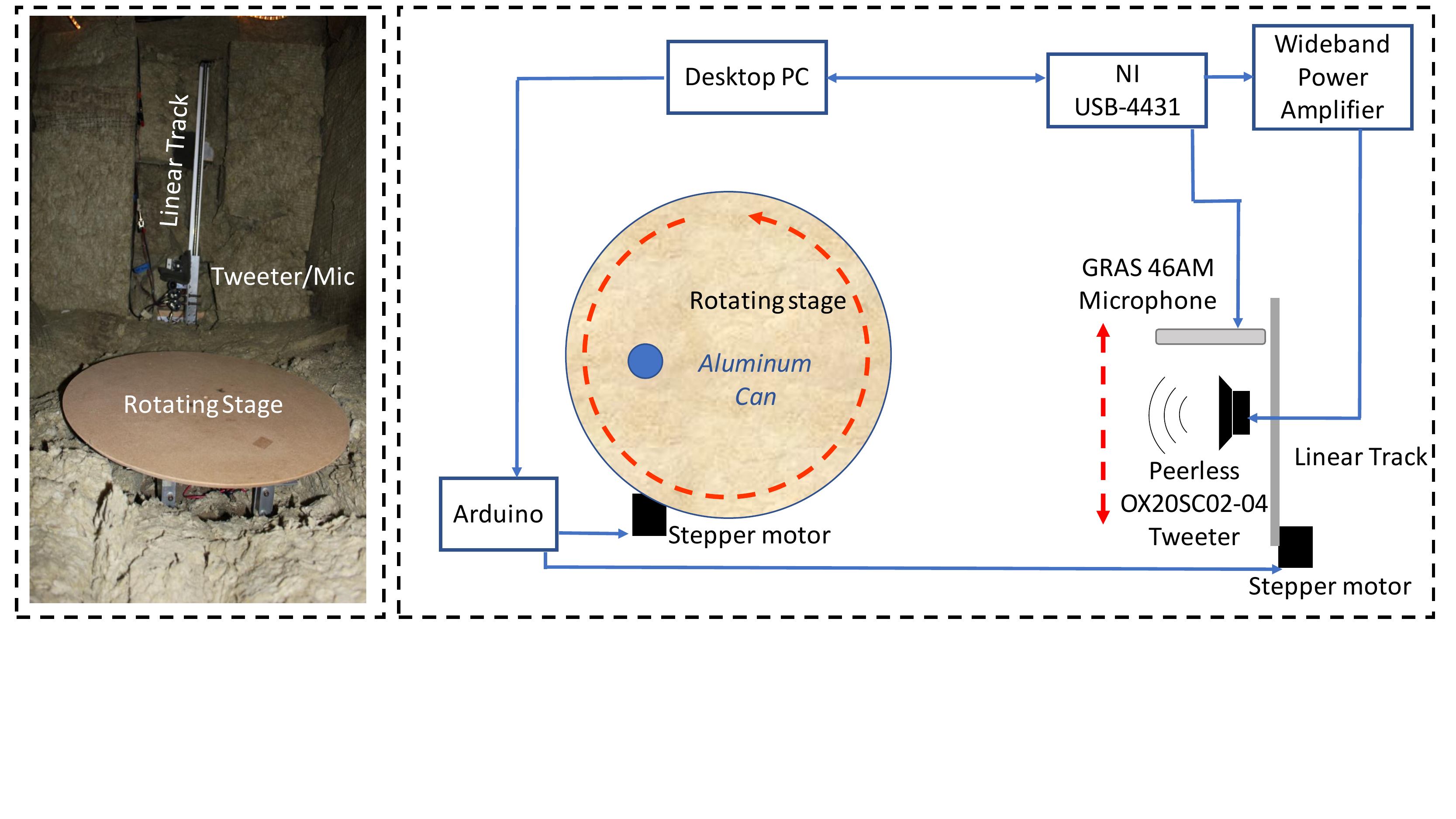}
\caption{Photo and block diagram of AirSAS. AirSAS consists of a tweeter and microphone directed at a rotating stage to mimic a circular SAS image geometry. The AirSAS system is setup in a confined area padded with sound-dampening material to suppress noise.} 
\label{fig:airsas_fig}
\end{figure*}

\textbf{Neural network deconvolution:} To perform SAS deconvolution, we leverage a neural network prior to implicitly regularize the scene reconstruction through an analysis-by-synthesis optimization. Figure \ref{fig:pipeline} illustrates our reconstruction pipeline. In the figure, the network is queried with a grid of coordinates and estimates five of the coordinates have relatively high amplitudes --- this network output is the deconvolved representation of the scene. We compute the complex-valued PSF by backprojecting simulated returns from a scatterer centered in the scene, and then convolve the network output with the PSF to obtain an estimated reconstruction of the scene. Our goal is to compare this estimate with the delay-and-sum reconstruction to compute a loss. As such, on the right side of the figure we illustrate the process of reconstructing the scene by delay-and-sum beamforming replica-correlated SAS measurements of the scene's backscattered acoustic energy. We then compare these two reconstructions to compute a loss that is backpropagated to the network weights. We repeat this process iteratively until the network learns a deconvolved representation of the SAS image.

 Formally, we model the scene as the output of a neural network that maps scene coordinates to a point scattering amplitude $F_{\theta} : (x, y) \mapsto \sigma(x, y)$ such that

\begin{equation}
    \begin{gathered}
    b_{\text{estimated}} = F_{\theta}*P_{\text{rc}},
    \end{gathered}
\end{equation}
where $\theta$ are the trainable parameters of the network. We solve for network parameters with the following optimization,

\begin{equation}
    \begin{gathered}
        \min_{\theta}||b_{\text{estimated}}(x, y) - \lambda(x, y)||_1,
    \end{gathered}
    \label{optim}
\end{equation}
where  $\lambda(x, y)$ are the delay-and-sum reconstructed amplitudes and $b_{\text{estimated}}(x, y)$ are the neural network's estimated amplitudes convolved with the scene PSF. We perform our loss using the complex representations of $b_{estimated}(x, y)$ and $\lambda(x, y)$ --- we find this provides sharper reconstructions compared with computing our loss using the absolute values of these signals.

\textbf{Neural network architecture:}
Our network architecture adopts the design presented in \cite{mildenhall2020nerf}. Our goal is to learn a non-linear function that maps input coordinates to their correct reflectivity amplitudes. As such, we construct a network architecture that serves as a template for this function --- the network weights define the function behavior. We learn the network weights through an analysis-by-synthesis optimization so that we do not require training data. Our network architecture consists of four fully connected layers, and each layer is followed by a non-linear activation function. The first three layers use ReLU activations to ensure we reconstruct positive amplitudes, and the final layer uses a Sigmoid activation to map amplitudes within $[0, 1]$. 

Recent work shows that neural networks exhibit a bias towards learning low spatial frequencies which diminishes reconstruction quality \cite{tancik2020fourier}. As such, we encourage our network to learn the scene's salient spatial frequencies by leveraging random Fourier features, as described in \cite{mildenhall2020nerf}. Formally, for each scene coordinate, $\mathbf{v} = (x,y)$, we compute $\gamma(\mathbf{v}) = [\cos{(2\pi \kappa \mathbf{B} \mathbf{v})}, \sin{(2\pi \kappa \mathbf{B} \mathbf{v})}]$, where $\cos$ and $\sin$ are performed element-wise; $\mathbf{B}$ is a vector randomly sampled from a Gaussian distribution $\mathcal{N}(0,I)$, and $\kappa$ is the bandwidth factor. Functionally, the bandwidth of the Fourier features regulates the cutoff for reconstructed spatial frequencies (i.e., increasing $\kappa$ allows for higher spatial frequencies). In practice, we choose a network bandwidth large enough to reconstruct salient scene features and ignore high-frequency noise features.


\begin{figure*}
\centering
\includegraphics[trim={0cm 11cm 0cm 0cm}, clip, width=\textwidth]{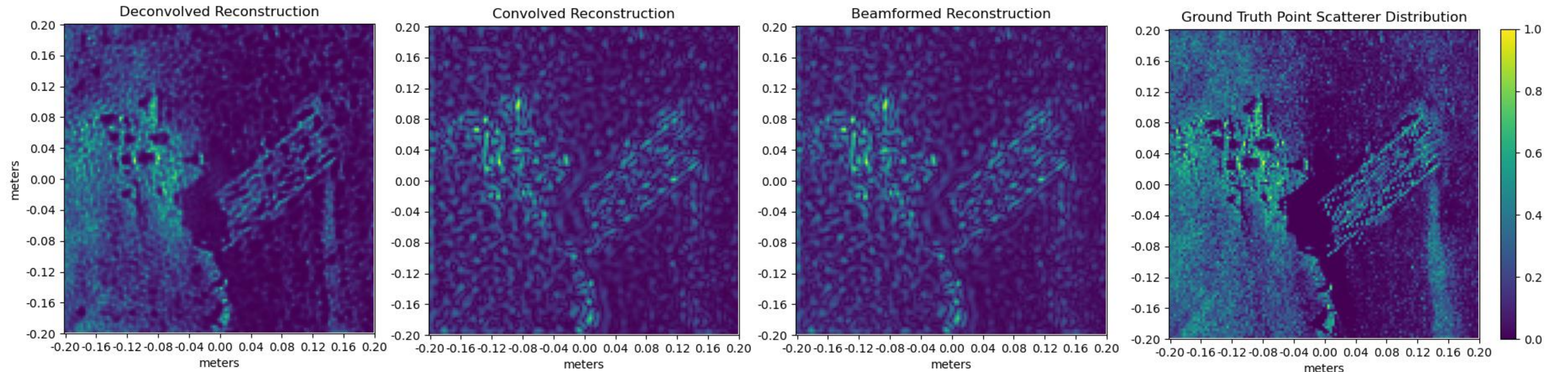}
\caption{Simulated results for our deconvolution method. The left-most image is the deconvolved output from the network, $F_{\theta}(x, y)$. The adjacent image is the convolved reconstruction obtained from convolving the network output with the scene PSF, termed $b_{\text{measured}}(x, y)$. The second to right-most image is the beamformed image $\lambda(x, y)$ of the ground truth scatterer position shown in the right-most image, $\sigma(x, y)$. We observe that our method removes the side lobe activity and closely matches the ground truth image. }
\label{fig:sim_results}
\end{figure*}

\begin{figure*}
\centering
\includegraphics[trim={0cm 0cm 0cm 0cm}, clip, width=0.90\textwidth]{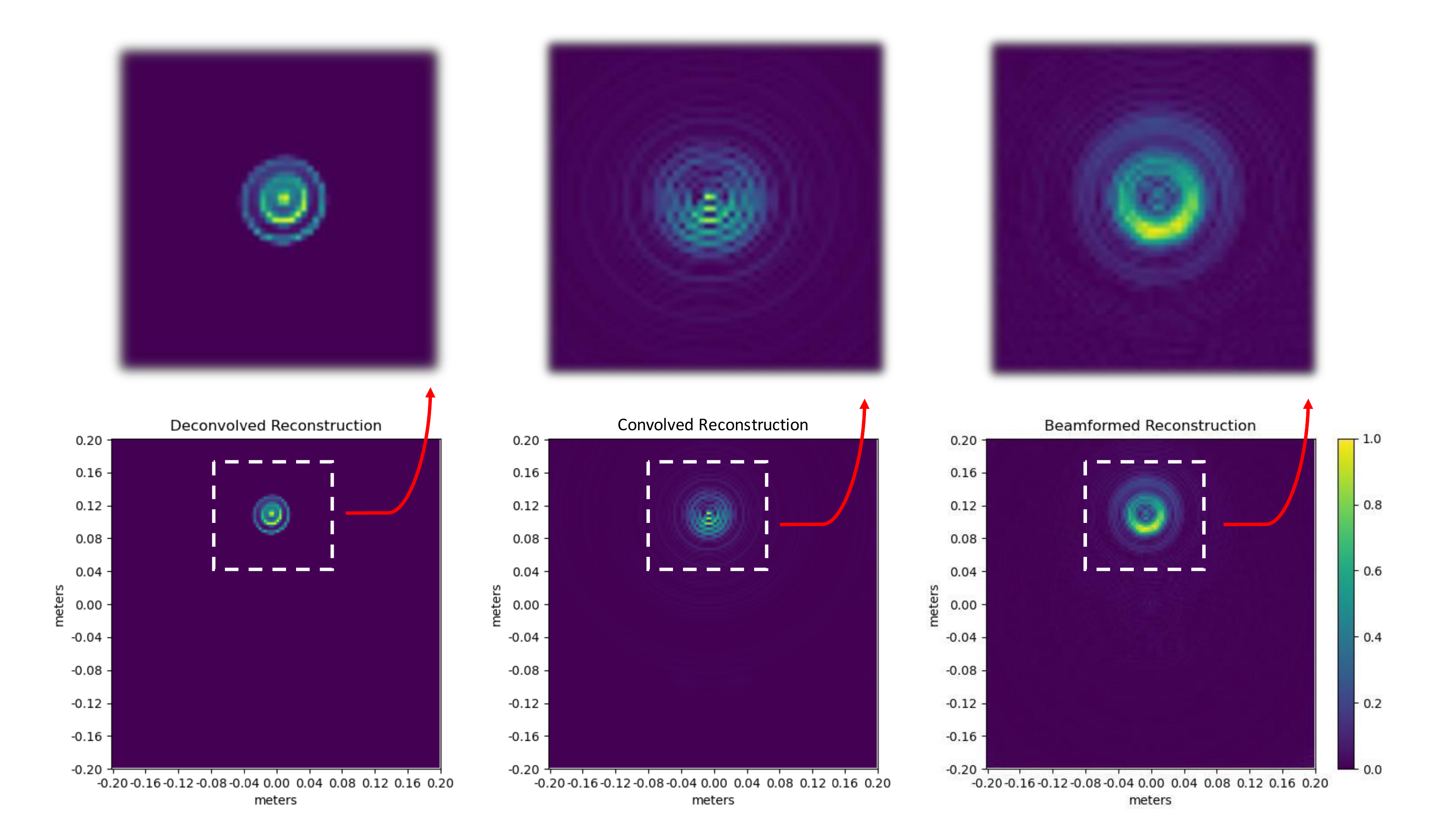}
\caption{This data is captured by imaging a aluminum can with AirSAS. The right-most image is the delay-and-sum beamformed reconstruction, the center image is our estimation of the beamformed reconstruction, and the leftmost image is our deconvolved output. We observe that our method reconstructs sharper boundaries for the aluminum can.} 
\label{fig:real_results}
\end{figure*}

\section{Implementation}
In this section, we describe the implementation for acquiring our simulated and real SAS measurements.

\textbf{Network implementation:} We optimize our network on a Tesla V100 GPU --- reconstructing a $129 \times 129$ scene takes approximately 5 minutes using a learning rate of $1e-4$ via Adam optimizer \cite{kingma2014adam}. We use $\kappa = 12$ and $\kappa = 20$ as the Fourier features bandwidth values for the real AirSAS and simulated reconstructions, respectively. 

\textbf{Simulated SAS measurements:} We simulate SAS measurements of a scene using a circular array geometry.  Specifically, we define a ring of $360$ transducers circling a $0.4 \times 0.4$ meter scene. We simulate the AirSAS geometry, where the array circles the scene $0.85$ meters radially from the center at a height of approximately $0.2$ meters. The scene is comprised of point scatterers with real-valued amplitudes --- we represent the point scatterers as the image titled \textit{Ground Truth Point Scatterer Position} in Figure \ref{fig:sim_results}. Each pixel in the $129 \times 129$ image represents a single point scatterer. We assume each scatterer is centered within its respective pixel and is amplitude modulated by the pixel intensity. To simulate SAS measurements for our geometry, we delay a LFM waveform (same parameters as AirSAS and described in next section) by the time-of-flight from the transducers to each scatterer in the image, and weight the delayed LFM by the amplitude of the scatterer. We then replica-correlate these waveforms and reconstruct the complex scene using coherent delay-and-sum beamforming. 

\textbf{AirSAS measurements:} In addition to simulated measurements, we capture in-air SAS measurements using AirSAS \cite{blanford2019development}. AirSAS consists of a tweeter and microphone directed towards a turntable. Rotating the turntable during the measurement process effectively mimics a circular SAS imaging geometry. We provide an illustration of the AirSAS setup in Figure \ref{fig:airsas_fig}. For our experiment, we place an empty aluminum can offset from the center of the platform and capture $360$ measurements (between 0 and 360 degrees) of the scene. We transmit a LFM waveform ($30$ kHz to $10$ kHz) for a duration of $.01$ seconds and modulated with a tapered cosine window (cosine fraction of $0.1$). We sample returns at $100$ kHz.

\section{Results} 
We provide an analysis of the results on our simulated data and AirSAS data. We note that we also compared our method to an off-the-shelf Wiener deconvolution \cite{van2014scikit}, but the results were exceptionally poor so we do not include them in the paper. All results are shown on a linear scale. 

\textbf{Simulated results:} In Figure \ref{fig:sim_results}, we show our method's ability to reconstruct fine details in an image that serves as our point scattering distribution. The ground-truth image, $\sigma(x, y)$ illustrates rocks and a man-made structure. We observe that these features become distorted in the beamformed reconstruction, $\lambda(x, y)$,  due to side lobe activity in the PSF. However, our method attenuates much of this side lobe activity through its optimization and reconstructs a more accurate representation, $F_{\theta}(x, y)$, of the original image. We observe that convolving the network output with the scene PSF yields a $b_{\text{measured}}(x, y)$ that is nearly indistinguishable from the beamformed reconstruction.

\textbf{AirSAS results:} We observe that beamformed reconstruction accurately captures the aluminum can geometry, but also contains blurring and side lobe activity. Specifically, the brightest ring in the beamformed reconstruction shows the outer edge of the aluminum can. This edge is blurred by the PSF, and a fainter ring forms around the outside of the can due to the PSF side lobes. We observe that our method sharpens the outer edge of the aluminum can substantially, providing clear reconstruction of its boundary. Further, our deconvolution method does not reconstruct the side lobe ring on the outside of the can. However, our method also reconstructs erroneous information within the boundary of the aluminum can. We believe this is due to a phase mismatch between the PSF and the real-world measurements. 

\section{Discussion}
In this work, we present a method for deconvolving SAS images using a neural network and analysis-by-synthesis optimization. We believe this work is an important first step for applying neural networks to SAS image deconvolution. However, there are several limitations to our current approach, and many research directions we intend to explore in the future. 

First, while our reconstruction of the simulated SAS measurements far outperforms delay-and-sum beamforming, these simulated measurements have some unrealistic properties. Specifically, we assume one point scatterer exists within the center of each image pixel. We delay-and-sum simulated measurements to the center of each pixel, meaning the reconstruction is perfectly in phase with the underlying scattering distribution. As such, our convolution of the network output with the complex scene PSF is phase aligned with the delay-and-sum beamformed image, a fact that will not hold true for real data. In real data, the point scatterer positions are not exactly known and far exceed the number of pixels resulting in the estimated PSF having a phase mismatch with the underlying scatterer distribution. We have observed in offline experiments that these phase mismatches can make the optimization unstable and sometimes not converge. While we are able to reconstruct the real AirSAS measurements (where the number of scatterers far exceeds the number of pixels since the underlying scattering distribution is continuous), our method does reconstruct erroneous features within the can edge, likely because of a phase mismatch. Future work should address the phase mismatch issue to ensure the optimization is more robust to a range of applications.

For future analysis and experiments, we wish to compare our deconvolution method to the suite of existing deconvolution methods tailored for optical applications. Additionally, we would like to better characterize our method with different array geometries and waveform bandwidths. Specifically, we wish to measure our reconstruction quality against a range of waveform bandwidths to better understand how our method applies to different hardware constraints. Further, the array geometry and its relation to the scene (e.g., if the array is undersampled) also affect the PSF geometry --- we believe future work should characterize the performance of our method in these cases.

\section{Acknowledgements}
This work was partially funded by ONR N00014-20-1-2330. The first author is funded by the DoD National Defense Science and Engineering Graduate Fellowship. The authors acknowledge Research Computing at Arizona State University for providing GPU resources that have contributed to the research results reported within this paper. 

\bibliographystyle{ieeetr}
{\footnotesize \bibliography{main}}

\begin{thebibliography}{10}

\bibitem{de1994deconvolution}
P.~de~Heering, K.~U. Simmer, E.~Ochieng-Ogolla, and A.~Wasiljeff, ``A
  deconvolution algorithm for broadband synthetic aperture data processing,''
  {\em IEEE Journal of Oceanic ngineering}, vol.~19, no.~1, pp.~73--83, 1994.

\bibitem{pailhas2017increasing}
Y.~Pailhas, Y.~Petillot, and B.~Mulgrew, ``Increasing circular synthetic
  aperture sonar resolution via adapted wave atoms deconvolution,'' {\em The
  Journal of the Acoustical Society of America}, vol.~141, no.~4,
  pp.~2623--2632, 2017.

\bibitem{hayes1992broad}
M.~P. Hayes and P.~T. Gough, ``Broad-band synthetic aperture sonar,'' {\em IEEE
  Journal of Oceanic Engineering}, vol.~17, no.~1, pp.~80--94, 1992.

\bibitem{ren2020neural}
D.~Ren, K.~Zhang, Q.~Wang, Q.~Hu, and W.~Zuo, ``Neural blind deconvolution
  using deep priors,'' in {\em Proceedings of the IEEE/CVF Conference on
  Computer Vision and Pattern Recognition}, pp.~3341--3350, 2020.

\bibitem{bretthorst1992bayesian}
G.~L. Bretthorst, ``Bayesian interpolation and deconvolution,'' tech. rep.,
  Washington Univ St Louis Mo Dept Of Chemistry, 1992.

\bibitem{sun2021coil}
Y.~Sun, J.~Liu, M.~Xie, B.~Wohlberg, and U.~S. Kamilov, ``Coil:
  Coordinate-based internal learning for imaging inverse problems,'' {\em arXiv
  preprint arXiv:2102.05181}, 2021.

\bibitem{mildenhall2020nerf}
B.~Mildenhall, P.~P. Srinivasan, M.~Tancik, J.~T. Barron, R.~Ramamoorthi, and
  R.~Ng, ``Nerf: Representing scenes as neural radiance fields for view
  synthesis,'' in {\em European Conference on Computer Vision}, pp.~405--421,
  Springer, 2020.

\bibitem{ulyanov2018deep}
D.~Ulyanov, A.~Vedaldi, and V.~Lempitsky, ``Deep image prior,'' in {\em
  Proceedings of the IEEE Conference on Computer Vision and Pattern
  Recognition}, pp.~9446--9454, 2018.

\bibitem{blanford2019development}
T.~E. Blanford, J.~D. McKay, D.~C. Brown, J.~D. Park, and S.~F. Johnson,
  ``Development of an in-air circular synthetic aperture sonar system as an
  educational tool,'' in {\em Proceedings of Meetings on Acoustics}, vol.~36,
  p.~070002, Acoustical Society of America, 2019.

\bibitem{gerg2020gpu}
I.~D. Gerg, D.~C. Brown, S.~G. Wagner, D.~Cook, B.~N. O'Donnell, T.~Benson, and
  T.~C. Montgomery, ``Gpu acceleration for synthetic aperture sonar image
  reconstruction,'' in {\em Global Oceans 2020: Singapore--US Gulf Coast},
  pp.~1--9, IEEE, 2020.

\bibitem{ferguson2005application}
B.~G. Ferguson and R.~J. Wyber, ``Application of acoustic reflection tomography
  to sonar imaging,'' {\em The Journal of the Acoustical Society of America},
  vol.~117, no.~5, pp.~2915--2928, 2005.

\bibitem{marston2011coherent}
T.~M. Marston, J.~L. Kennedy, and P.~L. Marston, ``Coherent and semi-coherent
  processing of limited-aperture circular synthetic aperture (csas) data,'' in
  {\em OCEANS'11 MTS/IEEE KONA}, pp.~1--6, IEEE, 2011.

\bibitem{park2020alternative}
J.~D. Park, T.~E. Blanford, D.~C. Brown, and D.~Plotnick, ``Alternative
  representations and object classification of circular synthetic aperture
  in-air acoustic data,'' {\em The Journal of the Acoustical Society of
  America}, vol.~148, no.~4, pp.~2661--2661, 2020.

\bibitem{richardson1972bayesian}
W.~H. Richardson, ``Bayesian-based iterative method of image restoration,''
  {\em JoSA}, vol.~62, no.~1, pp.~55--59, 1972.

\bibitem{lucy1974iterative}
L.~B. Lucy, ``An iterative technique for the rectification of observed
  distributions,'' {\em The Astronomical Journal}, vol.~79, p.~745, 1974.

\bibitem{daun2006deconvolution}
K.~J. Daun, K.~A. Thomson, F.~Liu, and G.~J. Smallwood, ``Deconvolution of
  axisymmetric flame properties using tikhonov regularization,'' {\em Applied
  Optics}, vol.~45, no.~19, pp.~4638--4646, 2006.

\bibitem{krishnan2011blind}
D.~Krishnan, T.~Tay, and R.~Fergus, ``Blind deconvolution using a normalized
  sparsity measure,'' in {\em CVPR 2011}, pp.~233--240, IEEE, 2011.

\bibitem{krishnan2009fast}
D.~Krishnan and R.~Fergus, ``Fast image deconvolution using hyper-laplacian
  priors,'' {\em Advances in Neural Information Processing Systems}, vol.~22,
  pp.~1033--1041, 2009.

\bibitem{xu2014deep}
L.~Xu, J.~S. Ren, C.~Liu, and J.~Jia, ``Deep convolutional neural network for
  image deconvolution,'' {\em Advances in neural information processing
  systems}, vol.~27, pp.~1790--1798, 2014.

\bibitem{park2020deformable}
K.~Park, U.~Sinha, J.~T. Barron, S.~Bouaziz, D.~B. Goldman, S.~M. Seitz, and
  R.~Martin-Brualla, ``Deformable neural radiance fields,'' {\em arXiv preprint
  arXiv:2011.12948}, 2020.

\bibitem{liu2020neural}
L.~Liu, J.~Gu, K.~Z. Lin, T.-S. Chua, and C.~Theobalt, ``Neural sparse voxel
  fields,'' {\em NeurIPS}, 2020.

\bibitem{reed2021dynamic}
A.~W. Reed, H.~Kim, R.~Anirudh, K.~A. Mohan, K.~Champley, J.~Kang, and
  S.~Jayasuriya, ``Dynamic ct reconstruction from limited views with implicit
  neural representations and parametric motion fields,'' {\em arXiv preprint
  arXiv:2104.11745}, 2021.

\bibitem{tancik2020fourier}
M.~Tancik, P.~P. Srinivasan, B.~Mildenhall, S.~Fridovich-Keil, N.~Raghavan,
  U.~Singhal, R.~Ramamoorthi, J.~T. Barron, and R.~Ng, ``Fourier features let
  networks learn high frequency functions in low dimensional domains,'' {\em
  NeurIPS}, 2020.

\bibitem{kingma2014adam}
D.~P. Kingma and J.~Ba, ``Adam: A method for stochastic optimization,'' {\em
  Proceedings of the 3rd International Conference on Learning Representations
  (ICLR)}, 2015.

\bibitem{van2014scikit}
S.~Van~der Walt, J.~L. Sch{\"o}nberger, J.~Nunez-Iglesias, F.~Boulogne, J.~D.
  Warner, N.~Yager, E.~Gouillart, and T.~Yu, ``scikit-image: image processing
  in python,'' {\em PeerJ}, vol.~2, p.~e453, 2014.

\end{thebibliography}

\end{document}